\documentclass[runningheads]{llncs}

\usepackage{eccv}

\usepackage{eccvabbrv}

\usepackage{graphicx}
\usepackage{booktabs}
\usepackage{subcaption}

\usepackage[accsupp]{axessibility}  
\usepackage{makecell}

\usepackage[pagebackref,breaklinks,colorlinks,citecolor=eccvblue]{hyperref}

\usepackage{orcidlink}

\begin{document}

\title{ActiveNeRF: Learning Accurate 3D Geometry by Active Pattern Projection} 


\author{Jianyu Tao*\inst{1}\and
Changping Hu*\inst{2} \and
Edward Yang\inst{3} \and Jing Xu\inst{2} \and Rui Chen \inst{2}}

\authorrunning{J.~Tao et al.}

\institute{
University of California, San Diego, CA 92093, United States. \and
Tsinghua University, Beijing 100084, China. \and Yale University, CT 06520, United States.}

\maketitle

\begin{abstract}
NeRFs have achieved incredible success in novel view synthesis. However, the accuracy of the implicit geometry is unsatisfactory because the passive static environmental illumination has low spatial frequency and cannot provide enough information for accurate geometry reconstruction. In this work, we propose ActiveNeRF, a 3D geometry reconstruction framework, which improves the geometry quality of NeRF by actively projecting patterns of high spatial frequency onto the scene using a projector which has a constant relative pose to the camera. We design a learnable active pattern rendering pipeline which jointly learns the scene geometry and the active pattern. We find that, by adding the active pattern and imposing its consistency across different views, our proposed method outperforms state of the art geometry reconstruction methods qualitatively and quantitatively in both simulation and real experiments. Code is avaliable at \url{https://github.com/hcp16/active_nerf}

\end{abstract}

\begin{figure*}
	\centering
	\includegraphics[width=\textwidth]{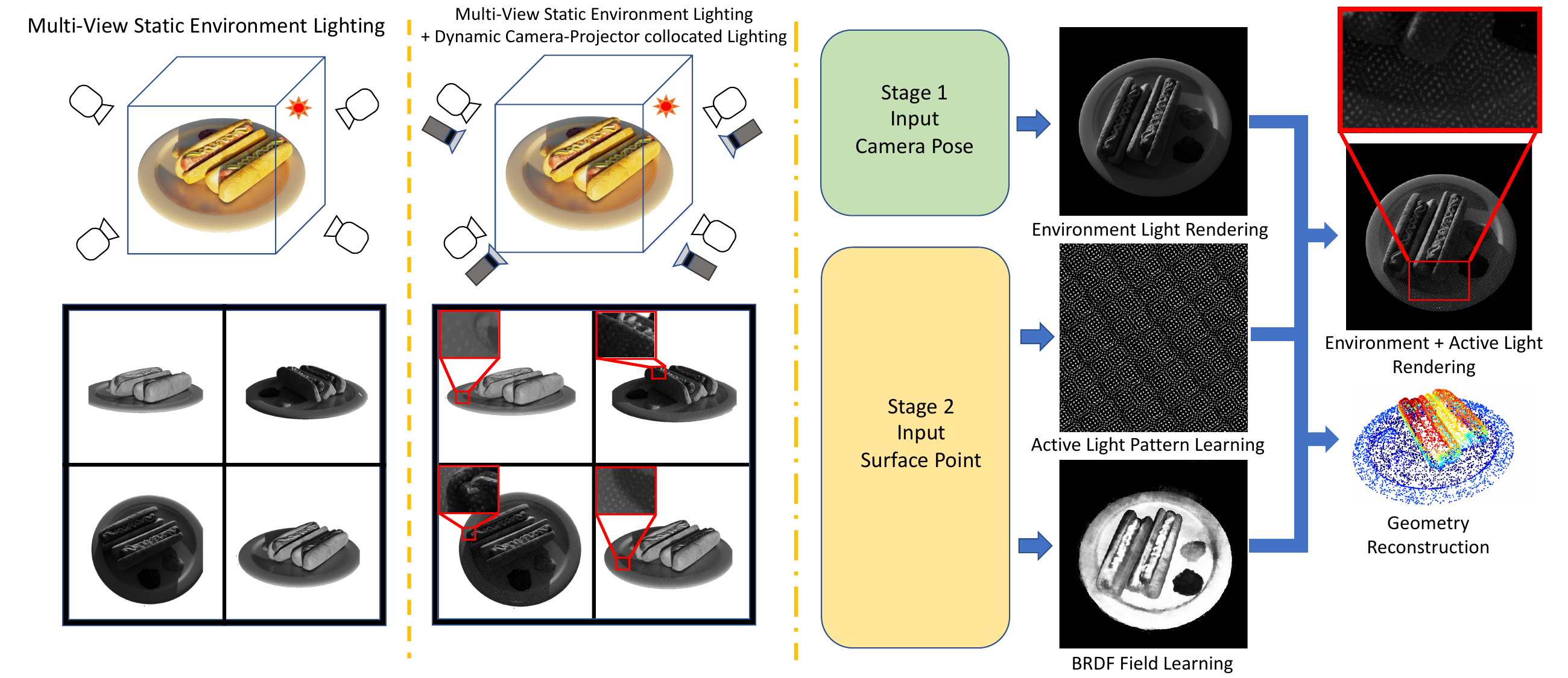}
	\caption{Different from the original NeRF setting where the environment lighting is static and has low spatial frequency, we use a projector to actively project patterns of high spatial frequency onto the scene. Our geometry reconstruction consists of two stages. The first stage only renders static environment light and outputs a rough geometry. the second stage utilizes the rough geometry to compute the active light intensity at each pixel, and fine-tunes the geometry and the active pattern jointly.}
	\label{fig:overview}
\end{figure*}

\section{Introduction}

Reconstructing geometry from multi-view images is an important and challenging problem in the computer vision community. Traditional methods~\cite{schoenberger2016mvs, openmvs2020} contain many hyperparameters and require hand-crafted features. Learning-based methods ~\cite{yao2018mvsnet, chen2019pointbased} are more robust to environment illumination and object texture and material but require a large-scale training dataset with ground-truth depth, which is costly and time-consuming to acquire in the real world. 

NeRF (Neural Radiance Field)~\cite{mildenhall2020nerf} is a new 3D representation which has shown success in many applications~\cite{ramamoorthi2023nerfs, pmlr-v205-kerr23a, ruckert2022neat}. NeRF does not require any additional supervision other than the images themselves, making it suitable for real-world applications. However, the geometry extracted from NeRF and its subsequent work is unsatisfactory. The reason is that these work assume a static passive environmental illumination that usually has low spatial frequency, limiting the amount of geometric information obtainable from the images.

In this paper, we propose ActiveNeRF, a novel approach that leverages the dynamic, high spatial frequency information provided by active pattern projection to improve the multi-view geometry reconstruction. Departing from the conventional NeRF setup, our method employs a pattern projector fixed to the camera to actively project a pattern onto the scene. The actively projected pattern augments the geometric information contained within the images due to its higher spatial frequency and controllable nature. Notably, this paradigm is readily implementable in real-world scenarios by utilizing an active stereovision depth sensor, such as an Intel RealSense D415.
The proposed setting deviates from the neural rendering equation assumption in NeRF, as the active light source introduces variability across images. To address this, we decompose the image into passive static environment light and active light components. This decomposition enables our rendering equation to accurately model the light changes across viewpoints.
Specifically, our method utilizes differentiable rendering through an implicit surface bidirectional reflectance distribution function (BRDF). The surface geometry and the active pattern are jointly optimized using the image synthesis loss. By imposing active pattern consistency across viewpoints, we can leverage the multi-view image information more effectively, leading to improved geometry reconstruction.

We first evaluate the performance of our proposed method on synthetic images of the same scenes as NeRF. In contrast to the original RGB images, we use grayscale images overlaid with active light patterns for our method. Despite the added complexity introduced by this altered problem setup, our method surpasses the current state-of-the-art NeRF2Mesh~\cite{tang2023delicate}, as well as traditional techniques such as OpenMVS~\cite{openmvs2020}. Furthermore, we evaluate the effectiveness of our method in real-world scenarios. Utilizing images captured from 24 viewpoints, it can accurately reconstruct the object geometry, significantly outperforming OpenMVS by a substantial margin.

In summary, our main contributions are: 
\begin{itemize}
	\item We propose a novel NeRF-based geometry reconstruction pipeline which only requires self-supervision on multi-view images, and achieves state-of-the-art performance both in simulation and the real world.
	\item We propose a learnable active light rendering framework that can jointly optimize the geometry and the active light pattern in an end-to-end self-supervised manner.
\end{itemize}

\section{Related Works}
\subsection{Multi-view 3D Reconstruction}
Given a set of input images from several viewing directions of a scene, multi-view stereo aims to reconstruct a 3D representation of the scene. Typically, this is done in four stages: cost computation or feature extraction, cost aggregation or matching features, optimization, and, finally, disparity refinement~\cite{988771}. In this setup, the intrinsic and extrinsic camera parameters for the images are assumed to be known or can be computed from the images themselves. The traditional multi-view 3D reconstruction pipeline estimates depth maps for each view which can then be fused into representations such as meshes, volumes, or point clouds.

Most methods that achieve state of the art results today are learning-based stereo methods that leverage large-scale benchmarks and higher computational ability to perform supervision~\cite{kitti, he2015deep, NIPS2012_c399862d}. Works such as PointMVSNet and MVSNet~\cite{yao2018mvsnet, chen2019pointbased, Mayer_2016} introduced end-to-end frameworks, allowing for all four stages of MVS to be done within a network. Recent unsupervised and self-supervised methods have achieved comparable results while also not needing ground truth depth. One such direction is image reconstruction loss, where an image's material and lighting are recovered along with its geometry~\cite{taniai2018neural, kaya2021uncalibrated}. Another such direction is reprojection which projects an image from one viewing angle to another and computes the difference between the actual image and the reprojected one~\cite{liu2021activezero, zhang2018activestereonet, app11125383}.

\subsection{Neural Radiance Fields}
Implicit neural representations~\cite{mildenhall2020nerf} are an alternative way of representing 3D objects that have seen significant advancement recently. NeRF and subsequent works use an MLP to represent a neural radiance field which takes in a 3D coordinate and view direction and outputs the volume density and radiance at that coordinate. Compared to traditional representations, implicit neural representations achieve better results on tasks such as novel view synthesis and are able to generate higher-fidelity renderings in general~\cite{sceneflow, park2021nerfies, zhang2022arf}. However, they struggle to predict geometry as there is no straightforward way to extract the scene surface from the output density. Dex-NeRF attempts to do this on transparent objects by truncating the weight along a ray with a pre-defined threshold and considering that as the surface point with limited success~\cite{IchnowskiAvigal2021DexNeRF}. Recent works such as VolSDF and NeuS~\cite{yariv2021volume, wang2023neus} aim to address this challenge by redefining the volume density using a signed distance function representation. Geo-NeuS improves on this by integrating sparse geometry and photometric consistency commonly found in multi-view stereo~\cite{fu2022geoneus}. NeRF2Mesh~\cite{tang2023delicate} takes a different approach to this task by decomposing environment light into specular and diffuse components before rendering, simplifying the information that the neural radiance field needs to learn.

Despite their success in predicting geometry, these methods assume environment lighting is static and has low spatial frequency such as a single circular light source projected onto the scene. This type of illumination imposes weak regularization on the implicit geometric information available for image synthesis. In contrast, we use active illumination, a setting where the active light pattern moves with the camera, resulting in the lighting of each image being different. Furthermore, this active light pattern has high spatial frequency, leading to stronger regularization overall. 

\section{Preliminary}
NeRF uses a multi-layer perceptron (MLP) to represent an implicit neural radiance field. The input to the model is a 5D vector containing a 3D position $\textbf{x}$, and 2D viewing direction $\textbf{d}_{out}$. In practice, $\textbf{d}_{out}$ is represented as a 3D unit vector. The output is a 4D vector containing a 3D RGB color vector $\textbf{c}$ and a scalar density $\sigma$. Thus, the MLP network with parameter $\theta$ can be denoted as $F_\theta:(\textbf{x},\textbf{d}_{out})\rightarrow(\textbf{c},\sigma)$. To render an image using NeRF, the camera's intrinsic and extrinsic parameters are used to compute the ray origin $\textbf{o}$ and ray direction $\textbf{d}_{out}$. The point along the ray can be computed by $\textbf{r}(t)=\textbf{o}+t\textbf{d}_{out}$. $t$ is bounded by $[t_{near}, t_{far}]$. The color of this camera ray can be represented as:
\begin{equation}
	C(\textbf{r})= \int_{t_{near}}^{t_{far}} T(t)\sigma(\textbf{r}(t))c(\textbf{r}(t),\textbf{d}_{out}) dt
\end{equation}
where:
\begin{equation}
	T(t)=\text{exp} \left( -\int_{t_{near}}^{t} \sigma(\textbf{r}(s))ds \right)
\end{equation}
$T(t)$ denotes the accumulated transmittance from $t_{near}$ to $t$, which can be interpreted as the probability that light travelling from $t_{near}$ to $t$ does not hit anything. In practice, NeRF samples points along the ray to approximate the volume rendering equation. Therefore, the original equation can be discretized to:
\begin{equation}
	C_{env}(\textbf{r})=\sum_{i=1}^N T_i(1-\text{exp} (-\sigma_i\delta_i))\textbf{c}_i
\end{equation}
where:
\begin{equation}
	T_i = \text{exp} \left(-\sum_{j=1}^{i-1}\sigma_j\delta_j \right)
\end{equation}

\section{Method}

\begin{figure*}[tb]
	\begin{center}
		\includegraphics*[width=\textwidth]{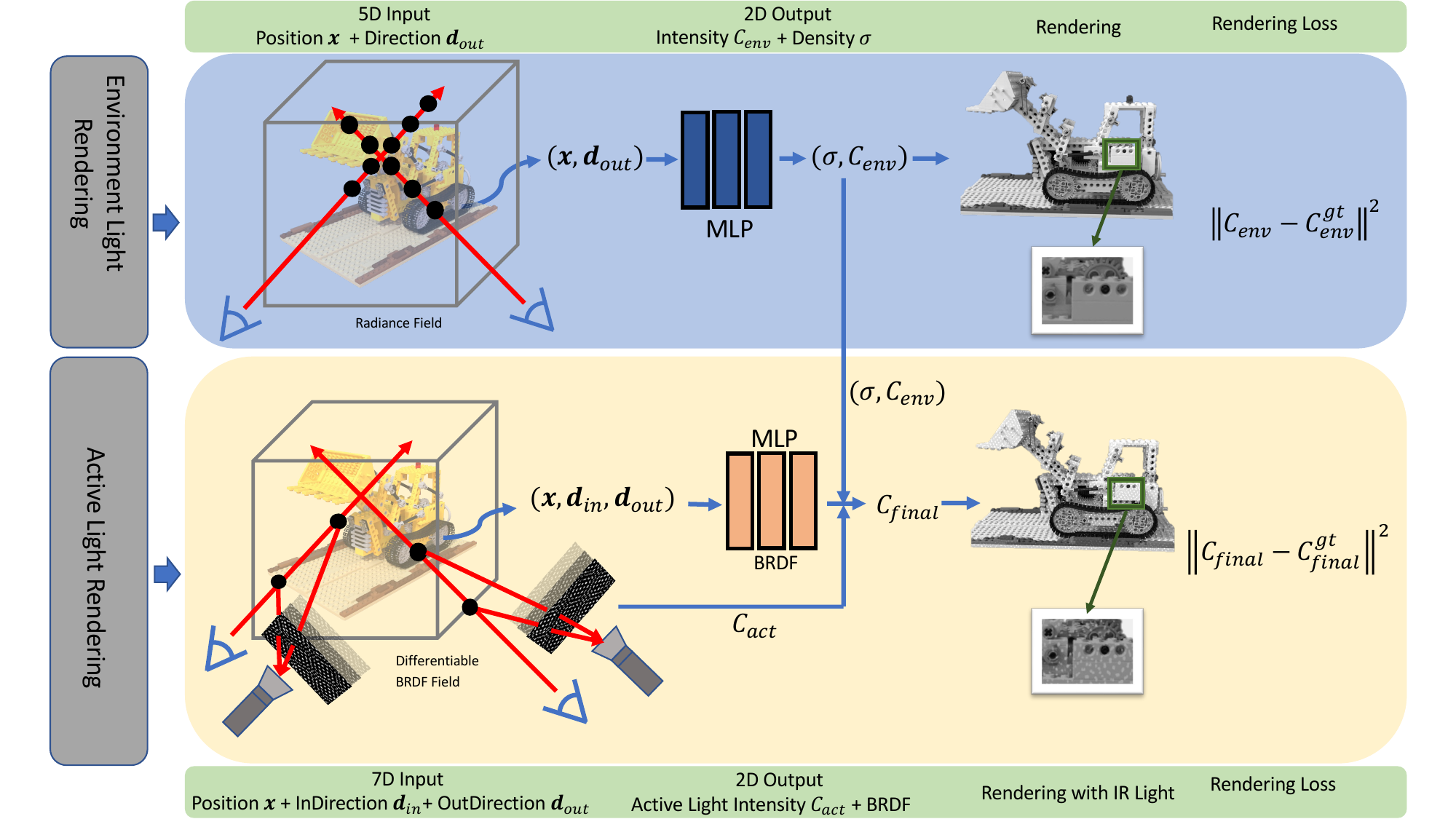}
	\end{center}    
	
	\caption{Architecture overview. The rendering process is divided into two stages. The first stage is similar to NeRF in that it renders images without active light (only environment light) and is supervised by ground truth images. The second stage renders active light onto the image from the first stage. Using the depth computed from the first stage, the second stage queries the BRDF value and active light radiance of a surface 3D point. This stage is supervised by ground truth images that contain an active light pattern. Since the whole rendering process is differentiable, the active light pattern is gradually learned from scratch.}
	\label{fig:architecture}
\end{figure*}

\subsection{Problem Statement}
We assume our environment only contains opaque objects such that the camera can capture the reflected active light in at least one view. And the object's surface can reflect some of the active light instead of absorbing most of it. We setup the camera and active light projector system in such a way that there is always a constant relative pose between them. This retains image synthesis quality while allowing our method to better understand geometry. We initially follow the classic NeRF setup but propose to synthesize novel view images with active light pattern, $I_{act}$, as well as images without, $I_{env}$.

\subsection{Method Overview}
Our task is to reconstruct geometry using a NeRF-like method in a scene containing active light, where the active light emitter's pose is fixed relative to the camera. We choose to use active light as opposed to other forms of illumination because the projected active light pattern will deform after hitting the object's surface. Furthermore, although the active light projector is moved with the camera, the active light pattern is still consistent across views. In this way, we can jointly optimize the active light pattern and the object geometry by rendering novel view images with active light. 

Similar to NeRF, we first query the network using the 3D position and viewing direction to get the environment radiance of the camera ray and a rough depth estimate. Then, we use the depth estimate to query the active light projector model for the active light radiance. Simultaneously, we initialize a 2D tensor to represent the active light pattern, $I_{pattern}$, mimicking an active light projector projecting the pattern onto an environment. We synthesize the final radiance by adding the environment radiance and the active light radiance together and then capping the value at 1. The whole process is differentiable, allowing the active light pattern to be updated throughout training. In this way, the model will learn to match the ground truth active light image while also implicitly regressing depth estimation more accurately. Intuitively this makes sense because the rendering process explicitly uses surface points to render active light: if the surface point is not accurate, the resulting rendered image's active light pattern will be shifted compared to the ground truth image's active light pattern. Our high-level framework is shown in Fig.\ref{fig:overview} and detailed description of the pipeline is shown in Fig.\ref{fig:architecture}

\subsection{Active Light Synthesis}
The volumetric rendering in NeRF assumes that environment light is static, so a point in 3D space and a 2D viewing direction can accurately determine the radiance. Thus, NeRF simplifies the volumetric rendering process to a setting where the objects themselves emit light. However, to render an image with an active light pattern coming from an active light projector moving with the camera, we need to trace the camera ray to the active light projector and compute the active light radiance. 

Here, we use a pinhole camera model to represent the active light projector, where $\mathbf{K}_{act}$ is the intrinsic matrix and $\mathbf{[R, t]}$ are the extrinsic parameters. Given a 3D surface point $\mathbf{x}$ in world coordinates, the pixel coordinate $\mathbf{p}$ can be computed by the intrinsic and extrinsic matrix.
Therefore, the radiance of a ray can be extracted from the active light pattern image. We then linearly interpolate the active light image to enable sub-pixel radiance computation. 

In summary, for each sampled point along the ray, the active light radiance is computed and added to the environment radiance. The resulting radiance will be used in the volumetric rendering equation to compute color:
\begin{equation}
	C_{final}(\textbf{r})=\sum_{i=1}^N T_i(1-\text{exp} (-\sigma_i\delta_i))(c_{env}^i + c_{act}^i)
\end{equation}

\subsection{Differentiable Surface BRDF}
Surface properties also play an important role in rendering active light images. For example, some surfaces are not lambertian, causing the active light to not reflect back to the camera at some viewing directions. To tackle this problem, we define a surface BRDF represented as an additional neural implicit field. The input to this field is a 9-D vector containing a 3D position vector$\mathbf{x}$, a 3D incident direction vector $\mathbf{d}_{in}$ and a 3D reflected direction $\mathbf{d}_{out}$. The output is a scalar representing the surface BRDF value $s$. In this way, the MLP network with parameter $\Theta_{act}$ can be denoted as $G_{\Theta_{act}}:(\mathbf{x},\mathbf{d}_{in},\mathbf{d}_{out})\rightarrow s$.

The final volumetric rendering equation we use is:
\begin{equation}
	\begin{aligned}
		& C_{final}(\textbf{r}) = \\ & \sum_{i=1}^N T_i(1-\text{exp} (-\sigma_i\delta_i))(c_{env}^i + G_{\Theta_{act}}(\textbf{x},\textbf{d}_{in},\textbf{d}_{out})c_{act}^i)
	\end{aligned}
\end{equation}

After we get the final radiance, we truncate this radiance so that the intensity is between 0 and 1. 

\subsection{Loss Formulation}
We supervise our method using images with, $C_{final}^{gt}$, and without, $C_{env}^{gt}$, active light pattern:
\begin{equation}
	L = \|C_{final} - C_{final}^{gt}\|^2 + \|C_{env} - C_{env}^{gt}\|^2
\end{equation}
\subsection{Two-Stage Training}

We find that simultaneously optimizing the model with both losses from the beginning leads to instability. This occurs when the predicted depths from NeRF in the early stage vary substantially across views, causing the gradients from pattern regularization to be ambiguous, resulting in a failure to converge. Therefore, when the coarse depth estimation converges, we start training both modules jointly to further refine the depth.

\subsection{Rendering Depth and Depth Fusion}
\begin{figure}[tb]
	\begin{center}
		\includegraphics*[width=0.8\textwidth]{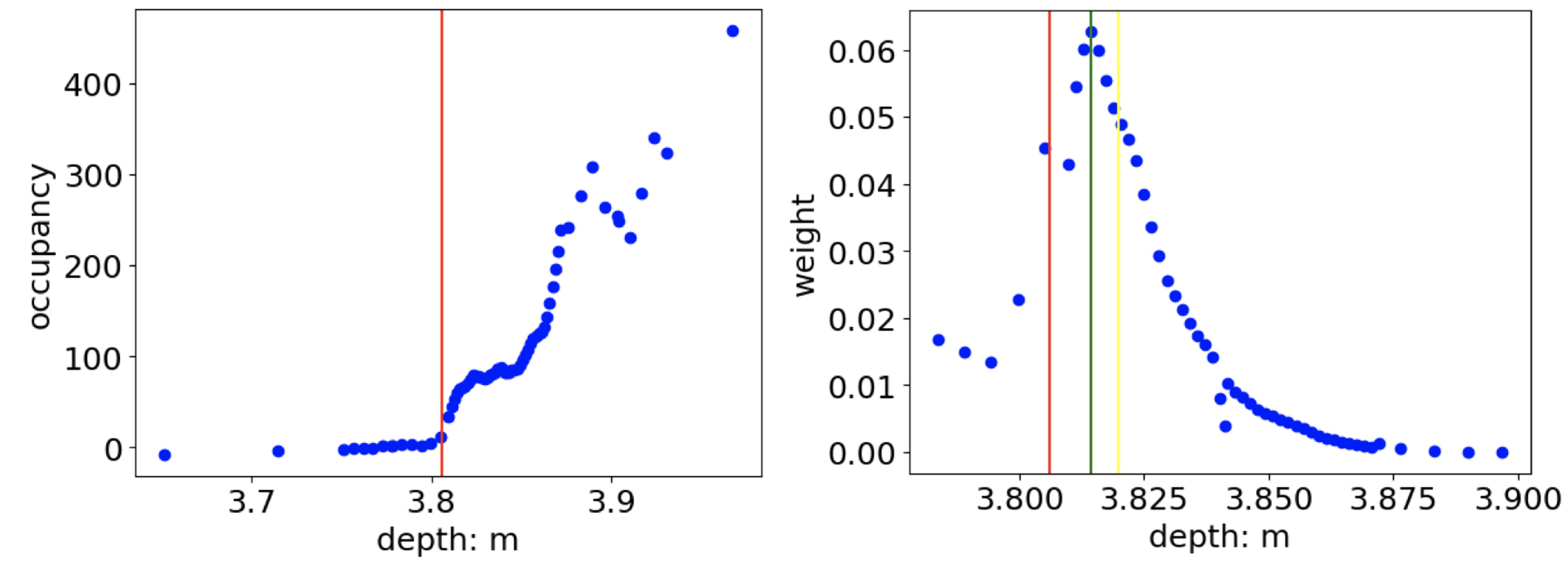}
	\end{center}    
        \vspace{-0.5cm}
	\caption{Visualization of occupancy and weight distribution along a single camera ray. The left figure is depth against occupancy $\sigma$ and the right figure is depth against weight $w$. The vertical red line denotes ground truth depth, the green line denotes weight-max depth, and the yellow line denotes weighted sum depth.}
	\label{fig:occuvisual}
\end{figure}

NeRF focuses on the rendering quality of novel view synthesis but does not explicitly regulate geometry. Although, the geometry can be implicitly estimated by the density from the neural radiance field:
\begin{equation}
	depth = \sum_{i=1}^n w_it_i
\end{equation}
where:
\begin{equation}
	w_i = T_i(1-\text{exp} (-\sigma_i\delta_i))
\end{equation}
$T_i$ corresponds to the depth value of each sampled point along the ray. Ideally, the weight of each sampled point that is not on the surface should be 0 and the color of points on the surface should be the final rendered color on that camera ray. In general, volumetric rendering should start from the object surface towards the observer. Thus, the weight at the surface should be the highest and decrease along the ray. However, in practice, the weight is non-zero along the ray. Usually, the density value increases near the surface and keeps increasing beneath the surface because the area below the surface is also occupied, as shown in Fig.\ref{fig:occuvisual}. Therefore, the surface computed by weighted sum over the samples' depth values will be below the ground truth object surface. 

We propose two ways to tackle this problem. First, following Dex-NeRF ~\cite{IchnowskiAvigal2021DexNeRF}, we can get a rough depth estimate by truncating the occupancy along the camera ray with a pre-defined threshold. This method requires tuning the threshold per scene which is not ideal. Second, we can render the depth directly using the point whose weight is the largest along the ray during inference time. In Fig.\ref{fig:occuvisual}, we show that the second option dramatically decrease the biases in depth estimation and does not require manual tuning for different scenes.

\section{Experiments}
\subsection{Dataset}
The dataset we use is a derivative of the original NeRF dataset. For each scene, we synthesize 100 images from different views. We use Blender~\cite{blender} to re-render the scene with a simulated active light projector whose position and orientation is constant relative to the camera. The resulting 800x800 image is 1-channel and gray-scale with values ranging from 0 to 1. The intrinsic and extrinsic matrices of the camera and active light projector are known for all the views. For each view, the dataset contains both the image with and without the active light projection.

\subsection{Implementation Details}
The environment radiance module, $C_{env}$, is modeled by an 8-layer MLP with 128 hidden size. The BRDF distribution module, $G_{\Theta_{act}}$, is modeled by a 6-layer MLP with 128 hidden size. Following NeRF, we also use positional encoding to improve the high-frequency surface performance in both modules. The position vector uses a 10-frequency positional encoding and the direction vector uses a 4-frequency positional encoding. 
We downsample the images to half resolution as the input. 
We train our model for 250k iterations within 6 hours on an NVIDIA RTX 3090. Our model is trained in an end-to-end fashion so the radiance field, BRDF field, and active light pattern are all trained at the same time. 

\subsection{Results}
\subsubsection{Geometry}
\begin{figure}[tb]
	\begin{center}
		\includegraphics*[width=0.7\textwidth]{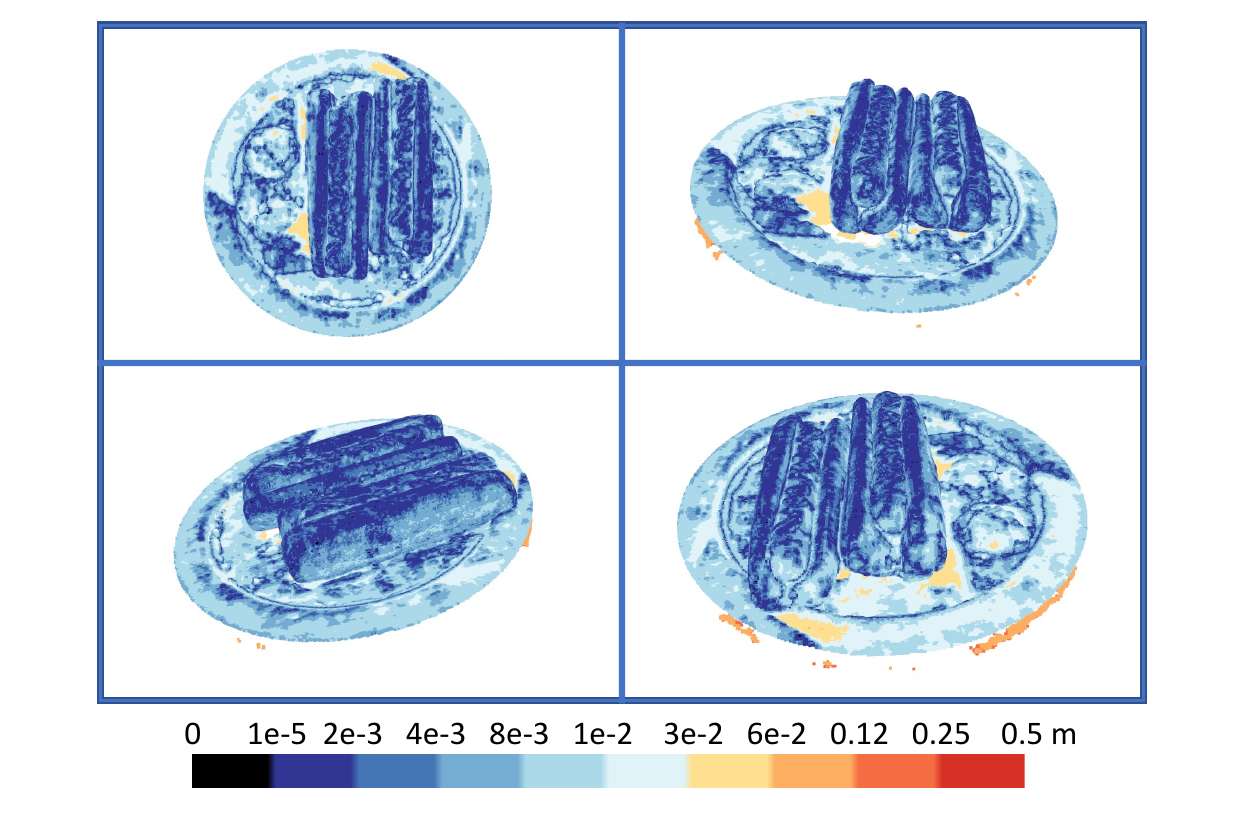}
	\end{center}
        \vspace{-0.8cm}
	\caption{Visualization of the reconstructed point cloud of synthetic images, colored with chamfer distance from the ground truth point cloud.}
	\label{fig:pcdvisual}
\end{figure}

To generate the ground truth point cloud, we use TSDF (Truncated Signed Distance Field) to fuse the depth maps for each view together to get the full reconstruction of the scene. We construct a 3D volume which covers the object in the scene to represent the geometry. The depth map for each view is then combined into the 3D volume to produce a TSDF volume. The final geometry will be extracted from this volume's zero surface. We represent our reconstruction as a point cloud so that the reconstruction quality can be evaluated using chamfer distance. In Fig.\ref{fig:pcdvisual}, we show the reconstruction quality of one scene from different viewing angles.  

\subsubsection{Active Light Pattern}
\begin{figure}[tb]
	\begin{center}
		\includegraphics*[width=\textwidth]{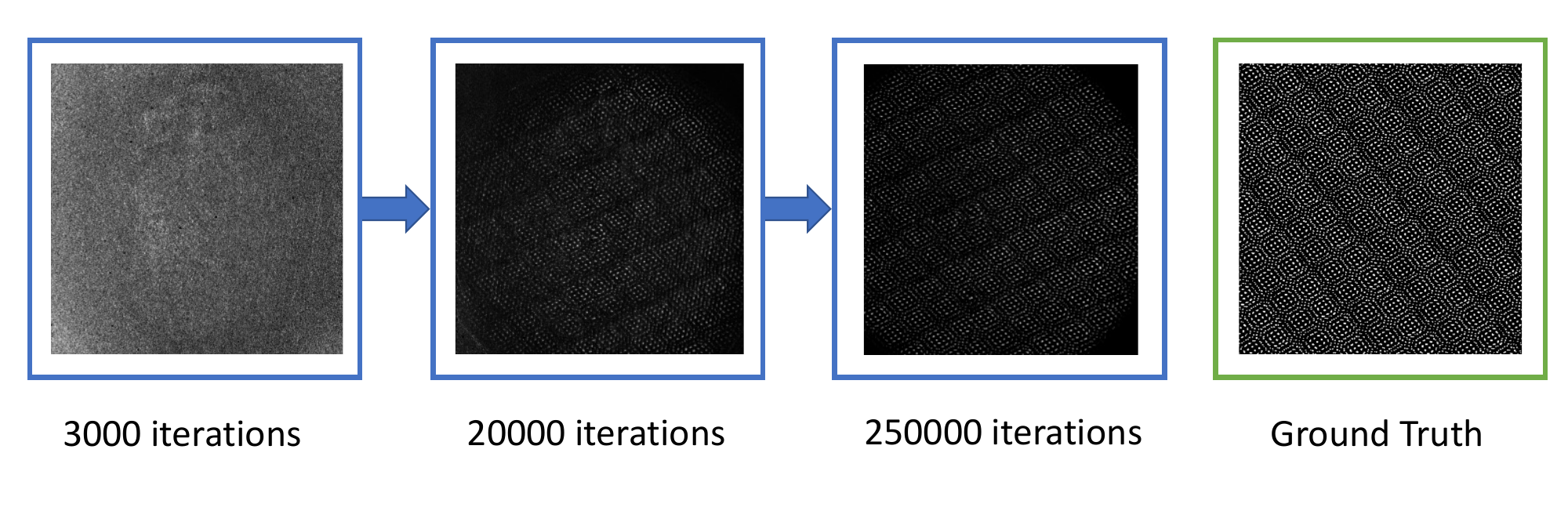}
	\end{center}    
        \vspace{-0.8cm}
	\caption{We observe that the active light tensor aligns more closely with ground truth as training progresses.}
	\label{fig:active lightPatternLearning}
\end{figure}

As shown in Fig.\ref{fig:active lightPatternLearning}, the active light pattern is represented as a 2D tensor which is updated throughout training via differentiable rendering. During inference time, this active light pattern is used to compute active light radiance in various positions and render the final image with active light.
\subsubsection{BRDF Field}

\begin{figure}[tb]
	\begin{center}
		\includegraphics*[width=0.75\textwidth]{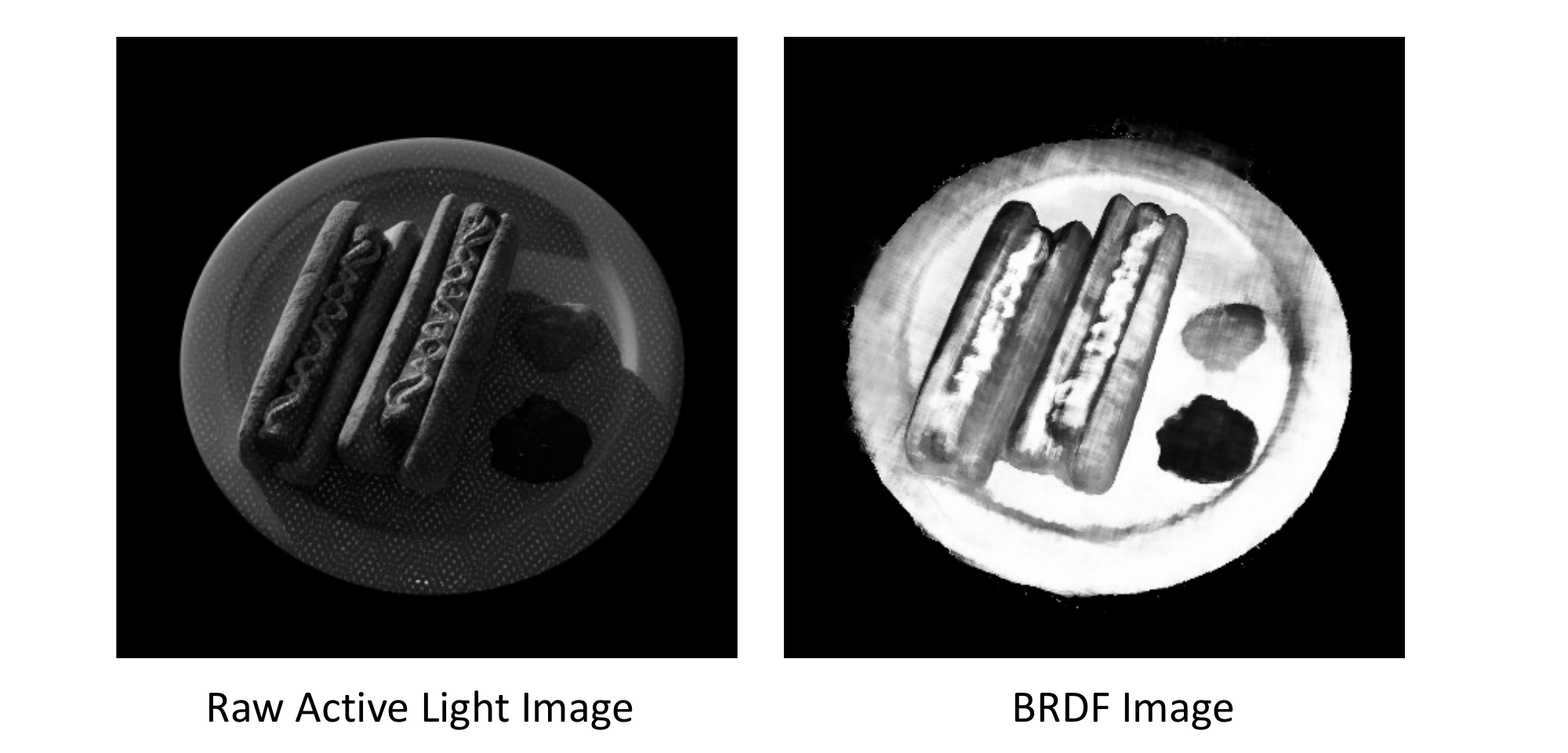}
	\end{center}    
        \vspace{-0.8cm}
	\caption{The BRDF field result in one sample view. Each pixel represents the BRDF value of the object surface at the light incident angle and exit angle computed by current camera position and active light projector position. }
	\label{fig:BRDFResult}
\end{figure}

As shown in Fig.\ref{fig:BRDFResult}, the active light radiance decreases in areas where BRDF is low/dark. Given a position, an input light direction, and an output direction, the MLP model will output a BRDF value. The positional encoding is also used here to learn high frequency material properties. 

\subsubsection{Active Light Rendering}

\begin{figure}[tb]
	\begin{center}
		\includegraphics*[width=0.6\textwidth]{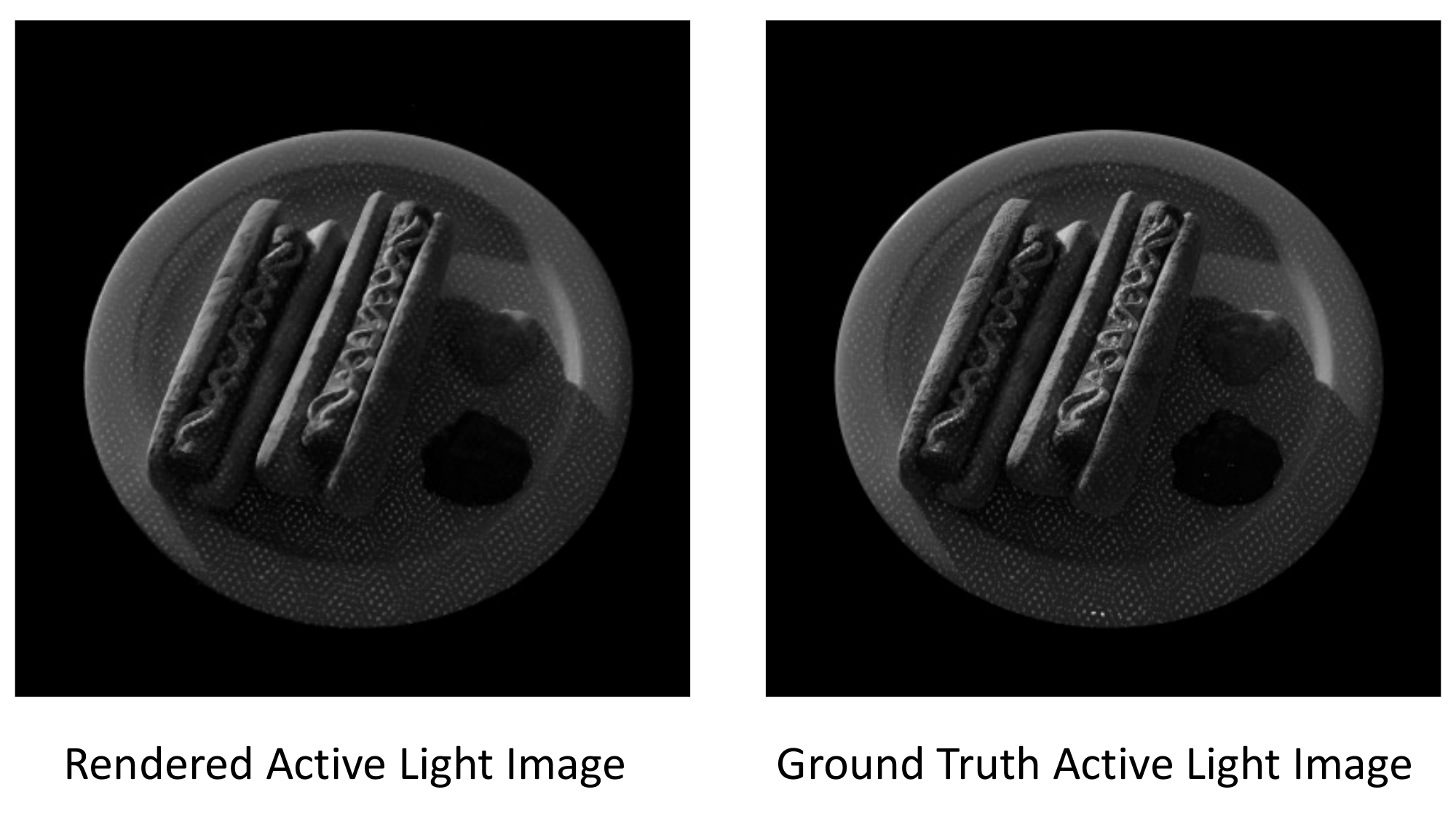}
	\end{center}    
 \vspace{-0.8cm}
	\caption{The final rendered image compared with the ground truth.}
	\label{fig:active lightImageResult}
\end{figure}

Given position, light incident direction,  and viewing direction, we use the trained BRDF and active light pattern to calculate the active light radiance. Then, as shown in Fig.\ref{fig:active lightImageResult}, we add the environment light radiance from the first stage to the active light radiance to render the final image. We evaluate the rendering quality using Peak Signal-to-Noise Ratio (PSNR). The average PSNR for synthesized images with active light pattern over all 8 scenes is 30.84. 

\subsection{Comparisons}
We use chamfer distance between our reconstruction and ground truth to evaluate our model performance. For both reconstruction and ground truth, we downsample the point clouds to a voxel size of 0.003 to ensure resolution consistency. To measure the reconstruction quality in more depth, we compute the percentage of chamfer distance that is lower than 0.01m and 0.05m. 

Since our method requires re-rendering the scene with active light, related methods which are evaluated on the original NeRF dataset cannot be easily adapted to our setting. We first compare our method with NeRF2Mesh~\cite{tang2023delicate}, which uses the same synthetic dataset as NeRF and directly outputs the geometry as a mesh. It is worth noting that NeRF2Mesh uses a 4-channel RGBA image as input where the exact foreground mask can be inferred directly from the alpha channel. Moreover, the meshing procedure removes outlier points. Although it is not a completely fair comparison, as shown in Tab.\ref{tab:comparison}, our method is superior to NeRF2Mesh in terms of the average chamfer distance and $<0.05$m inlier percentage. The results justify that our model can more accurately learn geometry by way of active pattern projection. 

Furthermore, we also compare our method to OpenMVS ~\cite{openmvs2020}. OpenMVS is a non-learning based algorithm, which reconstructs the geometry by detecting and matching correspondences across different views. As shown in Tab.\ref{tab:comparison}, our method also outperforms OpenMVS on most metrics.

\begin{table*}[tb]
	\centering
	\caption{Comparison of reconstruction accuracy across different methods}
	\begin{tabular}{l|llllllll|l}
		\hline
		&\multicolumn{8}{c}{OpenMVS (1-channel grayscale)} &\\ \hline
		& Chair &  Drums & Ficus & Hotdog & Lego & Materials & Mic & Ship & Mean\\
		\hline
		CD (mm  $\downarrow$) & N/A & 24.5 & 11.5 & 68.5 & 13.0 & N/A & 125.6 & 22.6 & 44.28 \\
		$P(CD<0.01\text{m}) (\%)$ ($\uparrow$) & N/A & 94.80 & 71.93 & 75.54 & 86.71 & N/A & 74.88 & 74.76 & 79.77 \\
		$P(CD<0.05\text{m}) (\%)$ ($\uparrow$) & N/A & 99.76 & 99.89 & 96.97 & 99.95 & N/A & 98.53 & 99.25 & 99.05 \\
		\hline
		&\multicolumn{8}{c}{OpenMVS (3-channel RGB)} &\\ \hline
		& Chair &  Drums & Ficus & Hotdog & Lego & Materials & Mic & Ship & Mean\\
		\hline
		CD (mm  $\downarrow$) & 20.6 & 28.7 & 13.3 & 30.4 & 13.8 & 27.4 & 420.2 & 19.0 & 71.7 \\
		$P(CD<0.01\text{m}) (\%)$ ($\uparrow$) & 92.07 & 96.08 & 76.89 & 77.71 & 88.54 & 84.81 & 49.36 & 68.01 & 79.20\\
		$P(CD<0.05\text{m}) (\%)$ ($\uparrow$) & 99.96 & 99.91 & 99.43 & 98.66 & 99.99 & 98.53 & 94.39 & 98.90 & 98.72\\
		\hline
		&\multicolumn{8}{c}{NeRF2Mesh (4-channel RGBA)} &\\ \hline
		CD (mm  $\downarrow$) & 6.5 & 86.3 & 15.3 & 38.2 & 17.8 & 15.4 & 13.8 & 98.4 & 36.4 \\
		$P(CD<0.01\text{m}) (\%)$ ($\uparrow$) & 94.94 & 55.94 & 91.10 & 49.51 & 80.78 & 73.24 & 87.90 & 31.83 & 70.65 \\
		$P(CD<0.05\text{m}) (\%)$ ($\uparrow$) & 99.97 & 75.56 & 96.46 & 72.00 & 94.24 & 98.74 & 99.52 & 55.96 & 86.56 \\
		\hline
		&\multicolumn{8}{c}{Ours (1-channel Grayscale) Truncated Occupancy Depth} &\\ \hline
		CD (mm  $\downarrow$) & 6.4 & 16.0 & 15.7 & 19.3 & 12.3 & 10.1 & 19.1 & 30.5 & \textbf{16.2} \\
		$P(CD<0.01\text{m}) (\%)$ ($\uparrow$) & 94.94 & 75.25 & 69.31 & 72.85 & 85.04 & 87.49 & 86.64 & 43.23 & 76.84 \\
		$P(CD<0.05\text{m}) (\%)$ ($\uparrow$) & 99.97 & 99.58 & 99.98 & 98.77 & 99.78 & 99.99 & 99.76 & 93.49 & 98.91 \\
		\hline
	\end{tabular}
	\label{tab:comparison}
\end{table*}

\begin{table*}[tb]
	\centering
        \caption{Comparison of reconstruction accuracy across different ablation experiments. }
	\begin{tabular}{l|llllllll|l}
		\hline
		CD (mm $\downarrow$)& Chair &  Drums & Ficus & Hotdog & Lego & Materials & Mic & Ship & Mean\\
		\hline
		w/o Active Light & 21.5 & 19.4 & 44.6 & 31.0 & 14.0 & 27.0 & 27.6 & 49.4 & 29.31 \\
		w/o BRDF& 83.9 & 47.9 & 42.8 & 335.9 & 13.7 & 42.8 & 152.2 & 107.1 & 103.29 \\
		{w/o Trunc. Occu. Depth}& 15.4 & 18.1 & 40.2 & 28.7 & 19.9 & 24.1 & 25.0 & 44.7 & 27.01 \\ \hline
  Ours & \textbf{6.4} & \textbf{16.0} & \textbf{15.7} & \textbf{19.3} & \textbf{12.3} & \textbf{10.1} & \textbf{19.1} & \textbf{30.5} & \textbf{16.2} \\
		\hline
		
	\end{tabular}
	\label{tab:ablation}
\end{table*}

\begin{table*}[tb]
	\centering
        \caption{Comparison of max weight depth, weighted sum depth and truncated occupancy depth. For each scene, the depth error is computed as the average across all of the views.}
	\begin{tabular}{l|llllllll|l}
		\hline
		Depth Error (mm $\downarrow$)& Chair &  Drums & Ficus & Hotdog & Lego & Materials & Mic & Ship & Mean\\
		\hline
		Weight Sum&  40.69 & 24.35 & 385.11 & 26.32 & 34.15 & 69.99 & 143.87 & 96.64 & 110.87 \\
		Weight-Max& 18.07  & 25.63 & 74.13 & 21.63 & 18.84 & 39.26 & 31.94 & 88.08 & 39.70 \\
		{Trunc. Occu. Depth  }& 14.51 & 30.86 & 77.57 & 17.04 & 21.85 & 14.61 & 26.53 & 49.75 & 31.59 \\
		\hline
		
	\end{tabular}
	\label{tab:weightmaxcomparison}
\end{table*}

\subsection{Ablation Studies}
In this section, we validate the effectiveness of each component and design choice through ablation experiments.

\subsubsection{w/o Active Light}
The main difference between NeRF and our method is that we not only render the static environment light but also the active light in novel view synthesis. We show that active light rendering can help the model regulate and refine the resulting depth. We compare the model performance with active light rendering and without. Functionally, the model without active light rendering is similar to NeRF. In Tab.\ref{tab:ablation}, we show that active light rendering greatly improves performance.

\subsubsection{w/o BRDF}

The BRDF module in our method serves as an auxiliary module that prevents non-lambertian surface properties, such as a specular surface, from hurting the performance of the active light rendering. As shown in Tab.\ref{tab:ablation}, this implicit BRDF field can improve both the rendering and reconstruction quality. This makes sense because non-lambertian surfaces don't reflect the active light pattern, resulting in an incorrectly learned active light pattern.

\subsubsection{Depth Weight Methods Comparison}

We evaluate various methods of extracting depth from the output density such as occupancy threshold (truncated occupancy depth) in Dex-NeRF ~\cite{IchnowskiAvigal2021DexNeRF}, max- weight depth (position of maximum weight along the ray), or weighted-sum depth (by weighted sum of weight and sampled points' depth along the ray). As shown in Tab.~\ref{tab:weightmaxcomparison}, truncated occupancy depth and max-weight depth can produce lower average absolute error across all views. This result supports our observation of the right-skewed weight distribution described in the Method section. 

\subsection{Baseline}

The baseline between the camera and the active light projector is crucial to the proposed method's performance. Since we query the active light intensity by triangulating surface points, the depth error's sensitivity is inversely proportional to the baseline. When the baseline is 0 and the camera and the active light projector are collocated with each other, the active light pattern on the object surface is not affected by geometry and the model cannot regulate the geometry by active light pattern. Therefore, we carry out more experiments in simulation on different baselines between camera and active light projector to verify this assumption. 

\begin{table}[h]
\centering
\caption{Comparison of reconstruction accuracy of different baselines between camera and active light projector. The scene used is Hotdog in NeRF~\cite{mildenhall2020nerf} synthetic dataset.}
\label{tab:baseline}
\begin{tabular}{l|lll}
    \hline
    Baseline & $\textbf{0.2m}$ (Ours) & $\textbf{0.4m}$ & $\textbf{0.6m}$ \\
     \hline
   $P(<0.01\text{m}) (\%)$ ($\uparrow$) & 72.85 & 92.19 & 92.70  \\
   $P(<0.05\text{m}) (\%)$ ($\uparrow$) & 98.77 & 99.71 & 99.89 \\
   Chamfer Distance (mm  $\downarrow$) & 19.3 & 10.6 & 10.2  \\
    \hline

\end{tabular}
\end{table}

As illustrated in Table \ref{tab:baseline}, increasing the baseline can further improve our method's performance. However, the marginal gain diminishes substantially with further increases. This observation aligns with the expected behavior, as once the ambiguity is largely resolved by increasing the baseline, additional increments yield diminishing performance improvement.

\subsection{Number of Views}
As our method leverages active pattern projection to improve geometry reconstruction, it can more effectively utilize the information contained within each view image. To validate the influence of view count on our approach, we evaluate our method on different numbers of views and compare it with NeRF.

\begin{figure}[tb]
	\includegraphics[width=\textwidth]{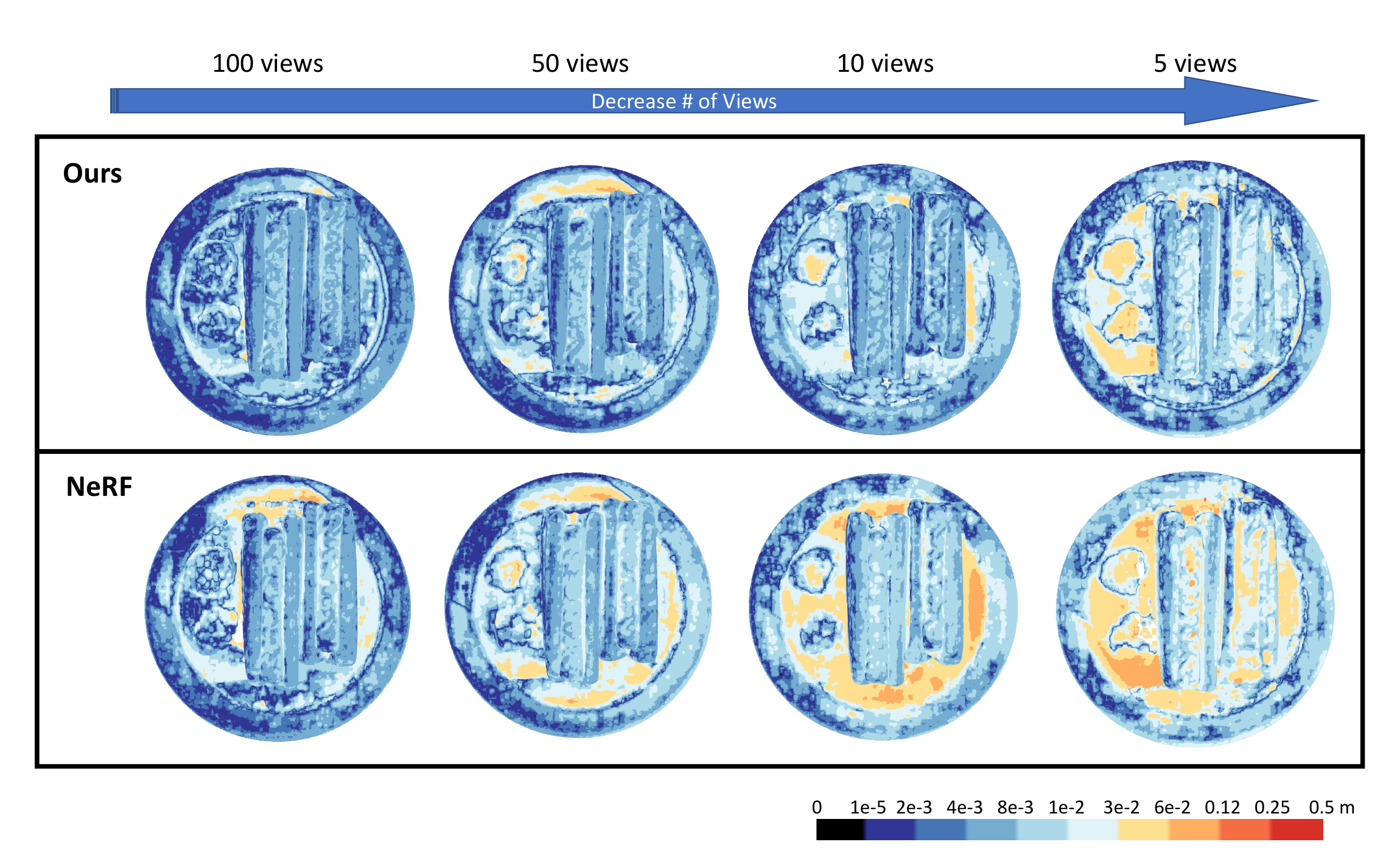}
  \caption{Comparison of reconstruction quality with different number of views.}
\label{fig:sparsecmp}
\end{figure}

We use a $0.4\textrm{m}$ baseline between the active light projector and the camera to achieve optimal active light regulation capacity. As illustrated in Table \ref{tab:numofviews} and Fig.~\ref{fig:sparsecmp}, decreasing the number of views degrades the performance of both methods.
Nevertheless, the degradation observed in our method is notably slower and less pronounced compared to NeRF. Remarkably, even with only 5 views, our method retains the capability to reconstruct high-fidelity geometry.
\begin{table}[h]
\centering
\caption{Comparison of reconstruction accuracy of different number of views available during training. The scene used is Hotdog in NeRF~\cite{mildenhall2020nerf} synthetic dataset.}
\label{tab:numofviews}
\begin{tabular}{l|llll}
    \hline
    &\multicolumn{4}{c}{NeRF}\\ 
    \hline
    Number of Views & \textbf{100} & \textbf{50} & \textbf{10} & \textbf{5} \\
     \hline
   $P(<0.01\text{m}) (\%)$ ($\uparrow$) & 65.98 & 61.13 & 52.92 & 35.67 \\
   $P(<0.05\text{m}) (\%)$ ($\uparrow$) & 98.53 & 97.89 & 94.29 & 90.17 \\
   Chamfer Distance (mm  $\downarrow$)    & 22.0 & 26.0 & 37.4 & 41.3 \\
    
    \hline
    &\multicolumn{4}{c}{Ours}\\ 
    \hline
    Number of Views & \textbf{100} & \textbf{50} & \textbf{10} & \textbf{5} \\
     \hline
   $P(<0.01\text{m}) (\%)$ ($\uparrow$) & 92.19 & 73.40 & 67.88 & 51.26 \\
   $P(<0.05\text{m}) (\%)$ ($\uparrow$) & 99.71 & 97.19 & 99.17 & 93.74 \\
   Chamfer Distance (mm  $\downarrow$)    & 10.6 & 19.2 & 20.5 & 30.0 \\
    \hline

\end{tabular}
\end{table}

\subsection{Real-world Experiments}

\begin{figure}[tb]
\begin{subfigure}{0.32\textwidth}
    \includegraphics[width=\textwidth]{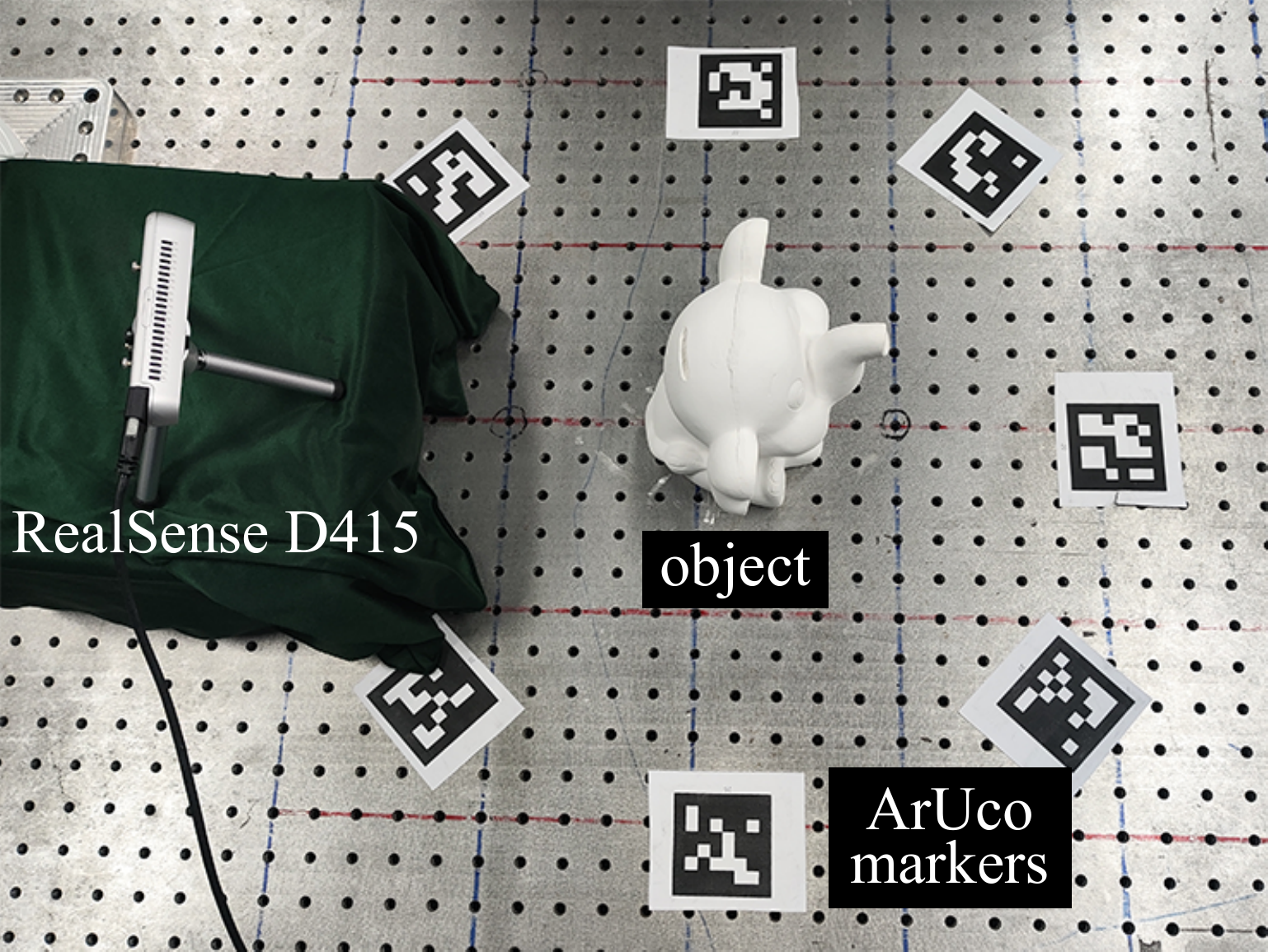}
    \caption{}
\end{subfigure}
\hfill
\begin{subfigure}{0.32\textwidth}
    \includegraphics[width=\textwidth]{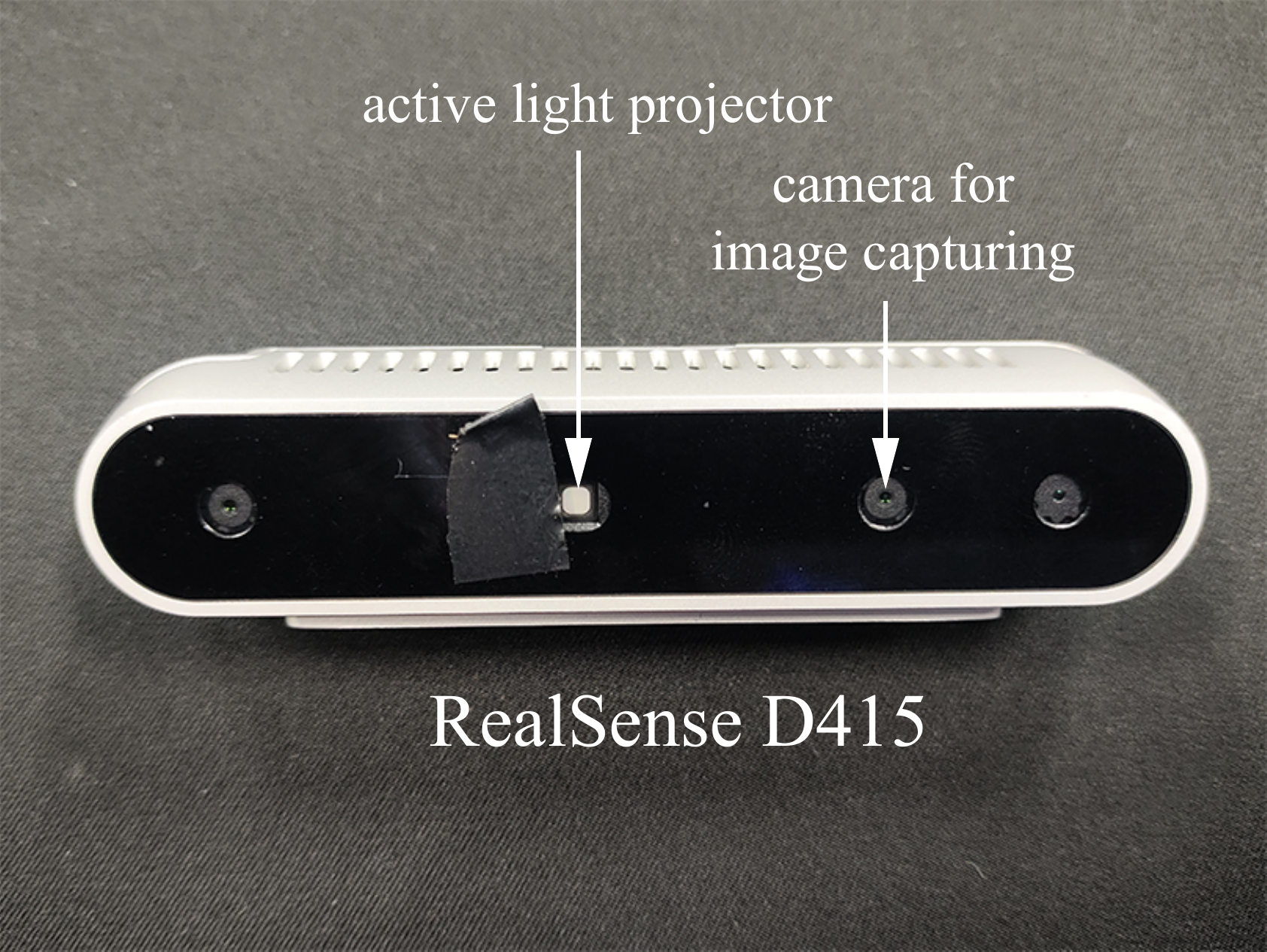}
    \caption{}
\end{subfigure}
\hfill
\begin{subfigure}{0.32\textwidth}
    \includegraphics[width=\textwidth]{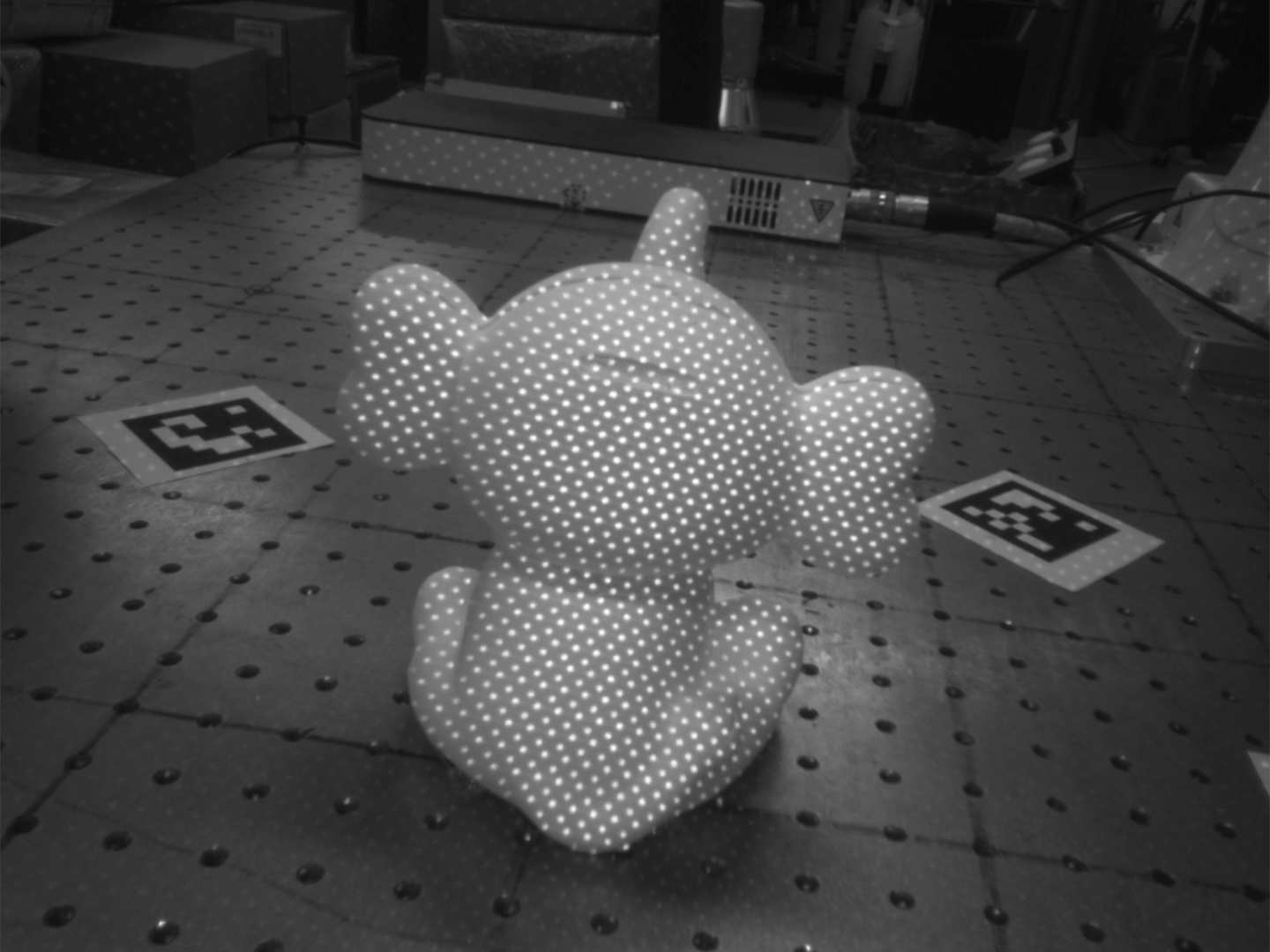}
    \caption{}
\end{subfigure}
\caption{(a) Real experiment setup; (b) RealSense D415 we use for image capturing; (c) Captured image with active pattern.}
\label{fig:real_exp}
\end{figure}

We further evaluate the effectiveness of the proposed method on real-world captured images. 
As shown in Fig.~\ref{fig:real_exp}, we employ a RealSense D415 to capture images with and without active pattern, such that the relative pose between the camera and projector remains constant during multi-view image capturing. 
Because the D415's pattern projector consists of 2 projectors, we cover one of the projectors.

\begin{figure}[tb]
	\begin{center}
		\includegraphics*[width=0.9\textwidth]{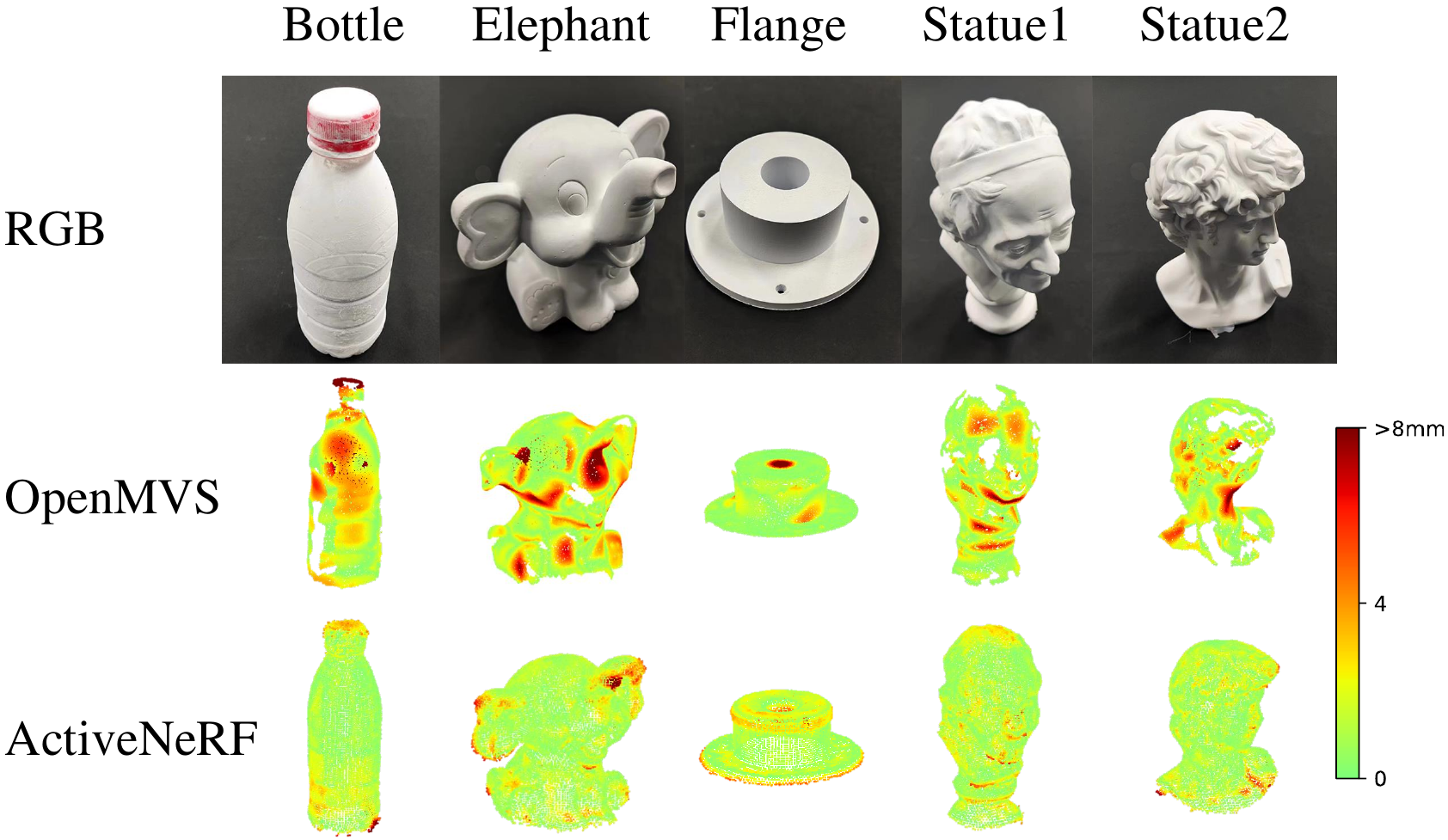}
	\end{center}
        \vspace{-0.4cm}
        \caption{RGB image and reconstructed point clouds. }
	\label{fig:real_results}
\end{figure}

We capture 5 objects at a distance of about 300~mm. 
For each object, we acquire images with $1920\times1080$ resolution from 24 views around it. 
At each viewpoint, images with and without active light are captured with the same exposure time. 
To obtain the extrinsic parameters for all the views, we place 8 ArUco markers around the object and utilize the marker detector in OpenCV to generate an initial estimation. Subsequently, we refine the extrinsic parameters by minimizing the reprojection loss. 
To obtain the relative position of the projector and the camera, we track the rays of some feature points in the pattern under the camera coordinate system and fit the intersection point as the projector position. 
For comparison, we also reconstruct the objects using OpenMVS~\cite{openmvs2020}, where we use the images without active pattern. 
We acquire the groundtruth shapes of objects using a handheld 3D laser scanner (CREAFORM HandySCAN 700), whose accuracy is 0.030~mm. 
We use chamfer distance between the reconstruction and the ground thruth to evaluate the performance. 
The results are shown in Table~\ref{tab:real_results} and Fig.\ref{fig:real_results}. 
Compared with OpenMVS, the point clouds reconstructed by our method are more accurate and with better coverage. 
OpenMVS performs worse in areas with less feature, such as the bottle body and the hair of the statues.

\begin{table}[tb]
    \centering
    \caption{Comparison of the proposed method and OpenMVS on the real data. }
    \label{tab:real_results}
    \begin{tabular}{l|lllll|l}
        \hline
          CD ($\text{mm} \downarrow$)  & Bottle & Elephant & Flange & Statue1 & Statue2 & Mean\\
        \hline
            OpenMVS~\cite{openmvs2020} & 5.57 & 5.84 & 6.56 & 5.74 & 7.11 & 6.61 \\
            Ours & \textbf{2.26} & \textbf{3.10} & \textbf{5.29} & \textbf{2.38} & \textbf{2.96} & \textbf{3.31} \\
        \hline
    \end{tabular}
\end{table}

\section{Conclusion and Limitations}
In conclusion, we present ActiveNeRF, a framework that can accurately reconstruct surface geometry from multi-view images with active pattern projection using only images with and without active light pattern for supervision. We find that, by jointly learning the geometry and active pattern, we can improve the reconstruction geometry quality. Experimental results demonstrate that our method outperforms state-of-the-art methods both on synthetic images and real captured images.

While our proposed method yields superior outcomes, there are certain limitations that require further investigation. The performance of our method relies on the visibility of the active light pattern, necessitating the assumption that the objects are diffuse and can effectively reflect the projected pattern. Consequently, reconstructing dark and transparent objects remains a challenging problem. Furthermore, our method assumes the object stays static during the multi-view capturing process. In future work, we aim to incorporate dynamic NeRF techniques to address more complex scenes encountered in real-world scenarios, enabling the reconstruction of dynamic objects.

\bibliographystyle{splncs04}
\bibliography{ref}
\end{document}